\documentclass{article} 
\usepackage{iclr2027_conference,times}

\iclrfinalcopy

\usepackage{amsmath,amsfonts,bm}

\def\eqref#1{equation~\ref{#1}}

\def\1{\bm{1}}

\def\vk{{\bm{k}}}

\def\vq{{\bm{q}}}

\def\vv{{\bm{v}}}

\DeclareMathAlphabet{\mathsfit}{\encodingdefault}{\sfdefault}{m}{sl}
\SetMathAlphabet{\mathsfit}{bold}{\encodingdefault}{\sfdefault}{bx}{n}

\usepackage{hyperref}
\usepackage{url}
\usepackage{graphicx}
\usepackage{wrapfig}
\usepackage{caption}   
\usepackage{amsthm}
\usepackage{booktabs}
\usepackage{array}
\usepackage{colortbl}
\usepackage{multirow}
\usepackage{enumitem}
\usepackage{textcomp}
 
\setlist[enumerate]{leftmargin=1.5em}
\setlist[itemize]{leftmargin=1.5em}
\hypersetup{hidelinks}

\title{CompKV: Compensation-Aware KV Selection for Long-Context LLM Inference}

\author{Zhen Huang\textsuperscript{1}\thanks{Equal contribution.}, 
Ruizhe Yao\textsuperscript{2}\footnotemark[1], 
Danyi Liu\textsuperscript{1}\footnotemark[1], 
Xinrui Chen\textsuperscript{1}, 
Shuwei Li\textsuperscript{1},
Siru Zhong\textsuperscript{1},
\\
\textbf{Zijian Cao\textsuperscript{3}, 
Yushan Lai\textsuperscript{1}, 
Mingming Guo\textsuperscript{1}, 
Weijie Zheng\textsuperscript{2}\thanks{Corresponding authors.}, 
Haohuan Fu\textsuperscript{1}\footnotemark[2]}
\\[0.4em]
\textsuperscript{1}Tsinghua University
\quad
\textsuperscript{2}Harbin Institute of Technology, Shenzhen
\quad
\textsuperscript{3}Northeastern University \\
\texttt{zhengweijie@hit.edu.cn} \quad \texttt{haohuan@tsinghua.edu.cn}
}

\begin{document}

\maketitle

\begin{abstract}
Despite their strong performance, large language models (LLMs) are bottlenecked by KV cache memory traffic during long-context inference. Sparse attention is widely used to accelerate LLM inference by computing exact attention over a selected subset of tokens. To recover the contribution of tokens excluded from exact attention, recent methods apply coarse-grained compensation to the omitted attention tail. However, existing methods typically select tokens based on attention mass and only then compensate for the unselected tokens. This decoupled design overlooks their interaction: selection should prioritize tokens that would leave the largest compensation error if omitted. To address this limitation, we introduce CompKV, the first compensation-aware sparse attention framework that divides tokens into blocks and explicitly optimizes selection for the downstream compensation mechanism. Our theoretical analysis shows that the residual left by block-level mean compensation is governed by both block attention mass and within-block logit variation. We approximate this residual using compact block-level statistics, yielding a deployable selection criterion. We further develop an efficient asynchronous implementation. Experiments on RULER and LongBench-Pro show that CompKV performs best among the evaluated sparse baselines while delivering up to a $6.85\times$ self-attention speedup over full attention.

\end{abstract}

\section{Introduction}
\label{sec:intro}

Large language models increasingly operate over long and evolving contexts for document understanding, software engineering, and agentic reasoning~\citep{llamateam2024llama,qwen3technicalreport,jimenez2024swebench,bai2024longbench,zhou2024webarena}. During autoregressive decoding, however, the KV cache grows with context length, and dense attention reads the entire cache at every step, making long-context inference increasingly bottlenecked by KV cache memory traffic.

To alleviate this bottleneck, query-aware sparse attention reduces this traffic by reading only a small set of tokens selected for the current query~\citep{xiao2024infllm,ribar2024sparq,wang2026fasa}. More recent methods divide tokens into blocks~\citep{tang2024quest,liu2025retrievalattention,deng2026unique} and further compensate for the attention contributions of omitted blocks using compact token summaries~\citep{hooper2025multipole,yang2026resa,fan2026flashprefillv2}. Despite recent progress in compensation, selection and reconstruction remain largely \textbf{decoupled}: blocks are ranked by predicted relevance or attention mass before reconstruction is considered. This design overlooks how well each block can be reconstructed by the compensator. A high-mass block may be easy to reconstruct, whereas a lower-mass block with greater within-block logit variation may leave a larger error. Therefore, selection should prioritize the blocks that have a larger estimated reconstruction error.

\begin{figure}[ht]
\centering
\includegraphics[width=\textwidth]{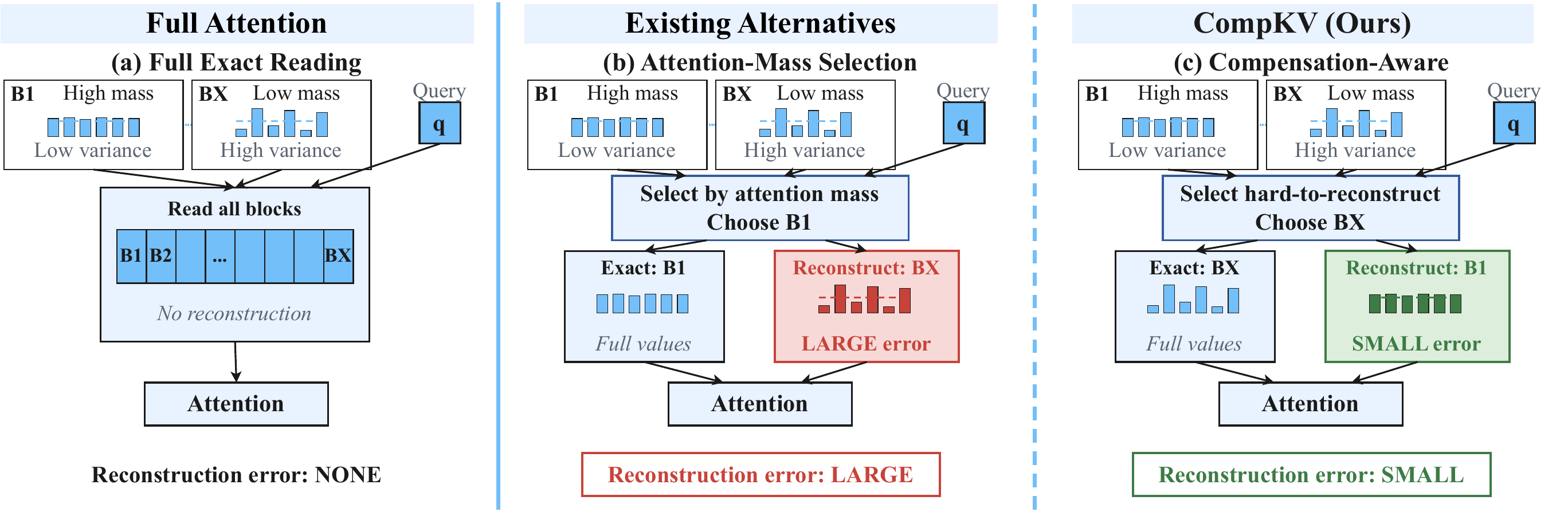}
\caption{Comparison between full attention (a), attention-mass selection (b), and CompKV (c) on two illustrative blocks B1 and BX. Mass-based selection may spend exact reads on blocks already well approximated, whereas CompKV targets blocks with the largest estimated reconstruction error.}
\label{fig:compkv-fig}
\end{figure}

Motivated by this mismatch, we introduce CompKV, a training-free, compensation-aware KV selection method that prioritizes blocks that are difficult to reconstruct, as shown in Figure~\ref{fig:compkv-fig}. To the best of our knowledge, CompKV is the first sparse attention framework to make block selection explicitly aware of the error induced by the downstream compensation mechanism. We formulate full block attention and its summary-based approximation as distributions whose KL divergence reduces to a log-partition gap. \textbf{For block-mean reconstruction, our analysis reveals that the residual contributed by an omitted block depends jointly on its attention mass and within-block logit variation.} Based on this result, CompKV estimates the post-compensation residual and then converts compensation-aware selection into a query-dependent block-ranking problem.


CompKV maintains compact mean and grouped-variance statistics for each KV block to estimate its post-compensation residual. For each query, these statistics approximate the block attention mass and within-block logit variation, which are combined into a selection score. Blocks with the largest scores receive exact attention, while the rest contribute through their mean summaries; both contributions are jointly normalized to produce the final output. For efficient long-context inference, we further develop an asynchronous CPU-offload implementation that overlaps summary transfer and compensation with block selection, CPU-side KV gathering, and selected-KV transfer.

We evaluate CompKV in terms of both accuracy and efficiency. For accuracy, we benchmark CompKV on RULER and LongBench-Pro using Llama-3.1-8B-Instruct, Qwen3-8B, and Qwen3-32B. Across both benchmarks, CompKV achieves the highest average scores among the evaluated sparse methods for all three models. For efficiency, we benchmark the CPU-offload single-layer pipeline on an NVIDIA H100 GPU. CompKV achieves the lowest mean decoding-step latency among the evaluated methods for all nine combinations of 32K-128K context lengths and 512-2,048 token budgets, delivering up to $6.85\times$ speedups over full attention.

Our contributions are:
\begin{itemize}
\item We identify the mismatch between selection and compensation and present the first formulation that explicitly models selection with respect to compensation. Our analysis shows that the post-compensation residual is jointly governed by attention mass and within-block logit variation.
\item Based on this analysis, we introduce CompKV, the first training-free, compensation-aware KV selection method that constructs a query-dependent score from compact block statistics, along with an efficient asynchronous implementation to accelerate long-context inference.
\item We comprehensively evaluate the accuracy and efficiency of CompKV. Across three widely used models and two long-context benchmarks, CompKV achieves the highest average scores among the evaluated baselines while delivering up to $6.85\times$ self-attention speedups over full attention.
\end{itemize}

\section{Related Work}

\textbf{KV Cache Reduction for Long-Context Inference.} To accelerate long-context inference, eviction methods bound the cache using recency, accumulated attention, heavy-hitter statistics, or prompt-side importance estimates~\citep{xiao2024efficient,liu2023scissorhands,zhang2023h2o,li2024snapkv,cai2025pyramidkv,feng2025adakv}. Complementary approaches reduce retained-state costs through low-rank compression~\citep{saxena2024eigen,singhania2024loki,chang2025palu}, quantization~\citep{liu2024kivi,hooper2024kvquant,kang2024gear,li2025kvtuner}, or memory management~\citep{kwon2023efficient,qu2025mobile}. These methods primarily reduce the amount or representation cost of retained KV states. In contrast, our work focuses on deciding which KV blocks should be read exactly for each query under a fixed per-step read budget.

\textbf{Query-Aware and Block-Level Sparse Attention.}
Query-aware sparse attention saves all KVs and conditions exact KV reads on the current decoding query. For example, Quest~\citep{tang2024quest} ranks pages with query-dependent upper bounds derived from compact key statistics, while InfiniGen~\citep{lee2024infinigen}, ShadowKV~\citep{sun2025shadowkv}, SpeCache~\citep{jie2025specache} and RetrievalAttention~\citep{liu2025retrievalattention} retrieve candidate blocks using different scoring mechanisms. However, these methods primarily rank blocks according to their predicted attention mass for the current query.

\textbf{Tail Compensation and Reconstruction.}
Tail-compensation methods approximate omitted KV states or their aggregate attention contribution. SparQ~\citep{ribar2024sparq} reallocates estimated omitted mass to a running mean value, RESA~\citep{yang2026resa} reconstructs the tail from a low-rank logit prior, and ResKV~\citep{zhan2026reskv} retains compact moment or residual summaries for compressed caches. These methods partly reconstruct the tail without extra exact KV reads.

Nevertheless, existing block selection and tail compensation are largely treated as separate stages, as mentioned in Section~\ref{sec:intro}. In contrast, CompKV explicitly selects blocks according to their estimated post-compensation residual, rather than simply those with the largest predicted attention mass.

\section{Compensation-Aware Selection Formulation}
\label{sec:formulation}

In this section, we formulate sparse attention selection and derive a compensation-aware selection criterion. We define Mean compensation as \textbf{replacing every logit in an unselected block by its block mean}. We use Mean compensation because its compact summaries yield an exact log-partition identity, making the interaction between selection and compensation analytically tractable. The compensation-aware formulation itself is not restricted to Mean compensation. Based on this, we analyze the residual introduced by compensation and show that the optimal selection priority depends jointly on a block's attention mass and the variation of its within-block attention logits.

\subsection{Mean Compensation Model and Selection Objective}
\label{sec:compkv-blocks}

\paragraph{Attention and Mean Compensation.} At each decoding step, we consider one KV head shared by $G$ query heads under grouped-query attention~\citep{ainslie2023gqa}. Let $\vq_g,\vk_j\in\mathbb R^d$ denote the query of head $g\in\{1,\ldots,G\}$ and the key at token position $j$, respectively, where $d$ is the head dimension. The corresponding attention logit is $z_{g,j}=\vq_g\vk_j^\top/\sqrt d$. Following~\citet{tang2024quest}, we divide the KV cache into contiguous blocks of a fixed size, starting from the first token. Let $\mathcal B$ denote the set of nonempty blocks, where each block $b\in\mathcal B$ contains $|b|$ tokens. Its unnormalized attention weight is $Z_{g,b}=\sum_{j\in b}e^{z_{g,j}}$, giving the full attention probability $P_g = \mathrm{softmax}_j(z_{g,j})$.

Given a selected block set $\mathcal S\subseteq\mathcal B$ shared by the $G$ query heads, for each token position $j\in b$, the logit under Mean compensation is
\begin{equation}
\hat z_{g,j}=
\begin{cases}
z_{g,j}, & b\in\mathcal S,\\
\bar z_{g,b}, & b\notin\mathcal S,
\end{cases}
\end{equation}
where $\bar z_{g,b}=|b|^{-1}\sum_{j\in b}z_{g,j}$ is the block-mean logit. Therefore, applying softmax jointly over all token positions defines the compensated distribution $\widehat P_g(\mathcal S)=\mathrm{softmax}_j(\hat z_{g,j})$.

\paragraph{Optimization objective.} For Mean compensation, we define the selection loss as $\mathcal L(\mathcal S)=\sum_{g=1}^{G}D_{\mathrm{KL}}\!\left(\widehat P_g(\mathcal S)\Vert P_g\right)$, summing over the query heads that share the selected set. With a budget of $K\le|\mathcal B|$ exact block reads, our objective is
\begin{equation}
\label{eq:gqa-exact-kl}
\min_{\mathcal S\subseteq\mathcal B,\,|\mathcal S|=K}
\mathcal L(\mathcal S).
\end{equation}


\subsection{Deriving the Compensation-Aware Selection Criterion}
\label{sec:compkv-residuals}

For each query head $g$, define the attention mass and logit variance of block $b$ as
\begin{equation}
\label{eq:compkv-mass-variance}
p_{g,b}=\frac{Z_{g,b}}{\displaystyle\sum_{c\in\mathcal B}Z_{g,c}},\qquad \sigma_{g,b}^2=\frac1{|b|}\sum_{j\in b}(z_{g,j}-\bar z_{g,b})^2,
\end{equation}
where $c$ ranges over all available blocks and $\sigma_{g,b}\ge0$ denotes the corresponding standard deviation.

To solve Equation~\ref{eq:gqa-exact-kl}, we expand the KL divergence defined in Section~\ref{sec:compkv-blocks}, yielding
\begin{equation}
\label{eq:mean-residual-objective}
\mathcal L(\mathcal S)
=
\sum_{g=1}^{G}
\log
\frac{\displaystyle\sum_{b\in\mathcal B}Z_{g,b}}
{\displaystyle
\sum_{b\in\mathcal S}Z_{g,b}
+
\sum_{b\notin\mathcal S}|b|e^{\bar z_{g,b}}
} = -\sum_{g=1}^{G}\log\left[1-\sum_{b\notin\mathcal S}p_{g,b}\left(1-\frac{|b|e^{\bar z_{g,b}}}{Z_{g,b}}\right)\right].
\end{equation}

Appendix~\ref{app:compkv-kl-proofs} provides the derivation. The ratio $|b|e^{\bar z_{g,b}}/Z_{g,b}$ measures the fraction of block $b$'s unnormalized attention weight recovered by Mean compensation. Appendix~\ref{app:exact-residual-selection} provides empirical support for this selection principle: ranking blocks by their exact compensation residuals reduces average attention KL and output error relative to ranking by exact attention mass.

A second-order Taylor expansion around the block-mean logit yields
\begin{equation}
\label{eq:mean-residual-expansion}
p_{g,b}\left(1-\frac{|b|e^{\bar z_{g,b}}}{Z_{g,b}}\right)=p_{g,b}(\frac12\sigma_{g,b}^2+O(\sigma_{g,b}^3)).
\end{equation}
Thus, the leading contribution of each block is proportional to its attention mass multiplied by its logit variance. Appendix~\ref{app:compkv-block-residual} derives this expansion and specifies the expansion regime.

Let $R_g(\mathcal S)$ denote the inner sum in
Equation~\ref{eq:mean-residual-objective}.
Applying $-\log(1-R_g)=R_g+O(R_g^2)$ to each head in
Equation~\ref{eq:mean-residual-objective} then yields
\begin{equation}
\label{eq:mean-residual-approximation}
\mathcal L(\mathcal S)=\frac12\sum_{g=1}^{G}\sum_{b\notin\mathcal S}p_{g,b}(\sigma_{g,b}^2+O\!\left(\sigma_{g,b}^3\right)).
\end{equation}
The remainder collects the blockwise expansion errors and the
outer-logarithm remainder, which is absorbed into the cubic term
in the small-variance regime.
Appendix~\ref{app:compkv-taylor} provides the derivation.

The second-order term is additive across unselected blocks. Since the sum over all blocks is independent of $\mathcal S$, minimizing the second-order term of Equation~\ref{eq:mean-residual-approximation} is equivalent to
\begin{equation}
\label{eq:mean-second-order-selection}
\max_{\mathcal S\subseteq\mathcal B,\,|\mathcal S|=K}\sum_{b\in\mathcal S}\sum_{g=1}^{G}p_{g,b}\sigma_{g,b}^2.
\end{equation}
\textbf{Thus, selecting the $K$ blocks with the largest $\sum_{g=1}^{G}p_{g,b}\sigma_{g,b}^2$ minimizes the second-order objective.} The resulting score combines a block's attention mass with the variation that Mean compensation leaves unreconstructed. A block with nearly constant logits contributes little to the loss, even when its attention mass is large. Appendix~\ref{app:compkv-drop} also proves that removing Mean compensation from our KL objective recovers the standard Top-$K$ selection rule based on attention mass for a single query head. Section~\ref{sec:compkv-score} estimates both factors from compact block statistics.

\section{CompKV}
\label{sec:compkv}

Building on the criterion derived in Section~\ref{sec:formulation}, CompKV turns the oracle selection rule into a practical decoding pipeline. It first estimates block attention mass and logit variance from compact statistics, uses their product to allocate exact reads, and represents the remaining blocks with mean summaries under a shared normalization. To realize it efficiently, we develop an implementation that overlaps compensation with KV retrieval and exact attention.

\subsection{Block Scoring from Compact Statistics}
\label{sec:compkv-score}

\paragraph{Grouped key-variance estimation.} Computing the exact score in Equation~\ref{eq:mean-second-order-selection} requires reading the keys of every candidate block. CompKV estimates this score from the mean key and grouped key variances stored for each block. The mean key $\bar{\vk}_b=|b|^{-1}\sum_{j\in b}\vk_j$ gives the block-mean logit $\bar z_{g,b}=\vq_g\bar{\vk}_b^\top/\sqrt d$ defined in Section~\ref{sec:compkv-blocks}. We partition the $d$ key coordinates into $r$ disjoint groups $\mathcal G_t$, indexed by $t=0,\ldots,r-1$. The vector $\bar{\mathbf D}_b\in\mathbb R^r$ stores the average coordinate variance in each group. At each decoding step, query head $g$ estimates the block's logit variance as
\begin{equation}
\label{eq:grouped-variance-metadata}
\widehat\sigma_{g,b}^2=\frac1d\sum_{t=0}^{r-1}\left(\sum_{i\in\mathcal G_t}q_{g,i}^2\right)(\bar{\mathbf D}_b)_t,\qquad
(\bar{\mathbf D}_b)_t=\frac1{|b|\,|\mathcal G_t|}\sum_{j\in b}\sum_{i\in\mathcal G_t}(k_{j,i}-\bar k_{b,i})^2.
\end{equation}
Here, $q_{g,i}$, $k_{j,i}$, and $\bar k_{b,i}$ denote the $i$th coordinates of $\vq_g$, $\vk_j$, and $\bar{\vk}_b$, respectively. The parameter $r$ controls the granularity of the stored variances. Appendix~\ref{app:compkv-grouped} specifies the coordinate groups and derives this estimate from the key covariance within a block along the query.

\begin{figure}[t]
\centering
\includegraphics[width=\textwidth]{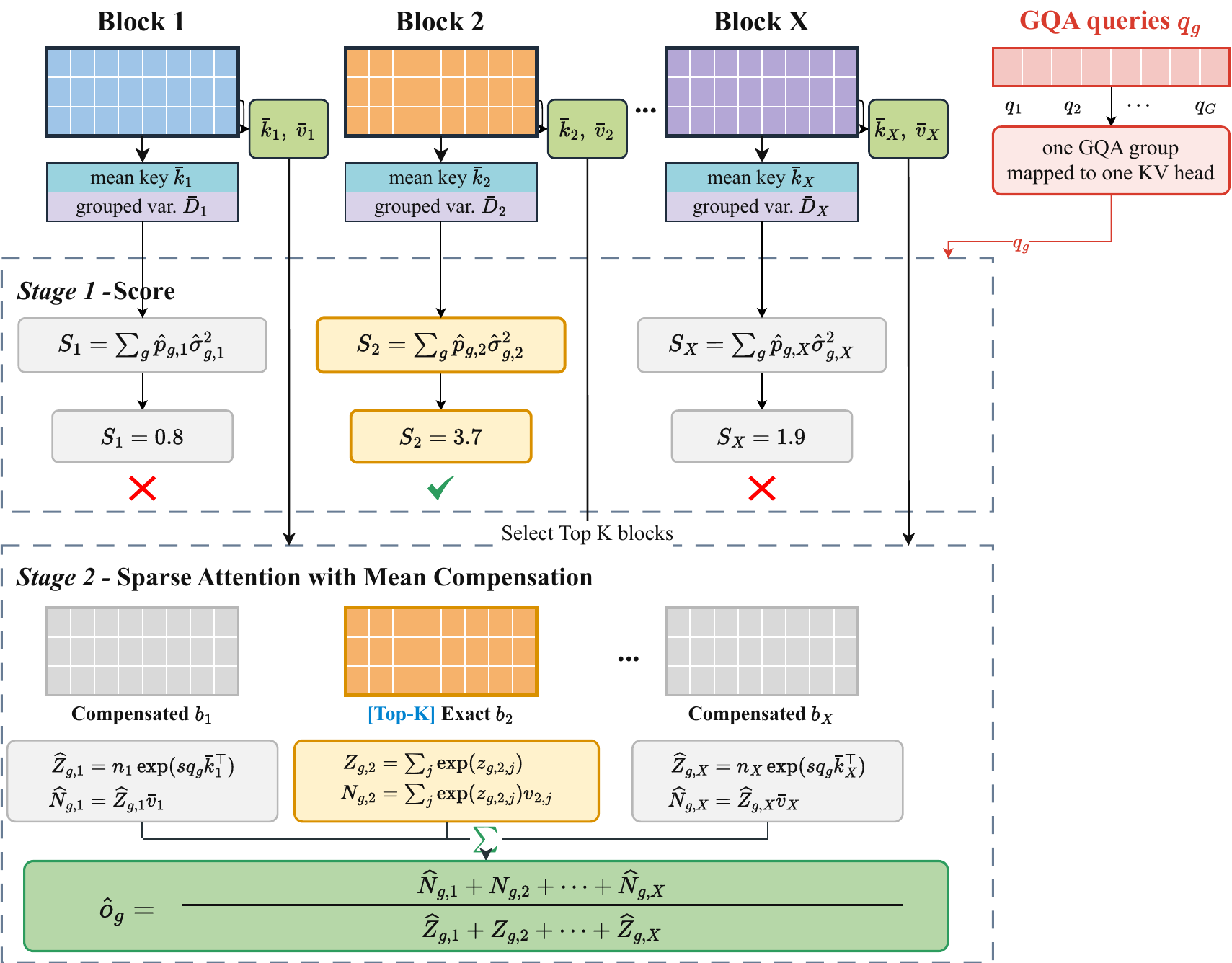}
\caption{End-to-end CompKV decoding, where $X=|\mathcal B|$ is the number of available blocks. Stage 1 ranks blocks with Equation~\ref{eq:compkv-score}. Stage 2 evaluates selected blocks exactly, compensates the remaining blocks with their means, and normalizes all contributions jointly.}
\label{fig:compkv-workflow}
\end{figure}

\paragraph{Attention-mass estimation.}
The log-partition expansion in Equation~\ref{eq:compkv-log-partition} gives $\log|b|+\bar z_{g,b}+\sigma_{g,b}^2/2$ as the second-order approximation to $\log Z_{g,b}$. Substituting $\widehat\sigma_{g,b}^2$ and normalizing the estimated block weights within each query head gives the attention-mass estimate $\widehat p_{g,b}$:
\begin{equation}
\label{eq:compkv-mass-estimate}
\widehat p_{g,b}=\frac{|b|\exp\!\left(\bar z_{g,b}+\frac12\widehat\sigma_{g,b}^2\right)}{\displaystyle\sum_{c\in\mathcal B}|c|\exp\!\left(\bar z_{g,c}+\frac12\widehat\sigma_{g,c}^2\right)}.
\end{equation}
The variance correction accounts for within-block logit variation when estimating the block's unnormalized attention weight. A detailed derivation is provided in Appendix~\ref{app:compkv-grouped}.

\paragraph{Compensation-aware block scoring.}
Replacing the mass and variance in Equation~\ref{eq:mean-second-order-selection} with these estimates defines the CompKV score $S_b$ for block $b$:
\begin{equation}
\label{eq:compkv-score}
S_b=\sum_{g=1}^{G}\widehat p_{g,b}\widehat\sigma_{g,b}^2.
\end{equation}

The two uses of $\widehat\sigma_{g,b}^2$ follow from the preceding derivations: the correction inside $\widehat p_{g,b}$ estimates the block's attention mass, and the outer variance factor captures the leading dependence of the compensation residual on logit variation. \textbf{CompKV computes $\widehat\sigma_{g,b}^2$ only once and then reuses it.} To evaluate the effectiveness of the score, we measure how closely it follows exact residual selection using Spearman correlation and residual capture. The results are shown in Appendix~\ref{app:practical-score-fidelity}.

\subsection{Block Selection and Compensated Decoding}
\label{sec:compkv-decode}

Figure~\ref{fig:compkv-workflow} shows the end-to-end decoding workflow of CompKV. Mean compensation additionally stores $\bar{\vv}_b=|b|^{-1}\sum_{j\in b}\vv_j$, where $\vv_j\in\mathbb R^d$. The complete summary $\{|b|,\bar{\vk}_b,\bar{\mathbf D}_b,\bar{\vv}_b\}$ contains $2d+r+1$ scalars per block and is updated incrementally as new K/V states enter the active block.

The mandatory set $\mathcal F$ contains the first sink block and the two most recent blocks, counting any overlap once. These blocks count toward the total budget, and Equation~\ref{eq:compkv-score} ranks the other blocks:
\begin{equation}
\label{eq:compkv-selected-set}
\mathcal S=\mathcal F\cup\operatorname{TopK}_{K-|\mathcal F|}\left\{S_b:b\in\mathcal B\setminus\mathcal F\right\},\qquad|\mathcal S|=K.
\end{equation}

Here, $\operatorname{TopK}$ returns the blocks with the highest scores. Selected blocks use their original K/V states, while unselected blocks use their mean summaries. Both contributions are normalized together:
\begin{equation}
\label{eq:compkv-output}
\widehat{\mathbf o}_g=\frac{\displaystyle\sum_{b\in\mathcal S}\sum_{j\in b}e^{z_{g,j}}\vv_j+\sum_{b\notin\mathcal S}|b|e^{\bar z_{g,b}}\bar{\vv}_b}{\displaystyle\sum_{b\in\mathcal S}Z_{g,b}+\sum_{b\notin\mathcal S}|b|e^{\bar z_{g,b}}}.
\end{equation}

\subsection{Efficient Implementation of CompKV}
\label{sec:efficient-implementation}

\begin{figure*}[t]
  \centering
  \includegraphics[width=\linewidth]{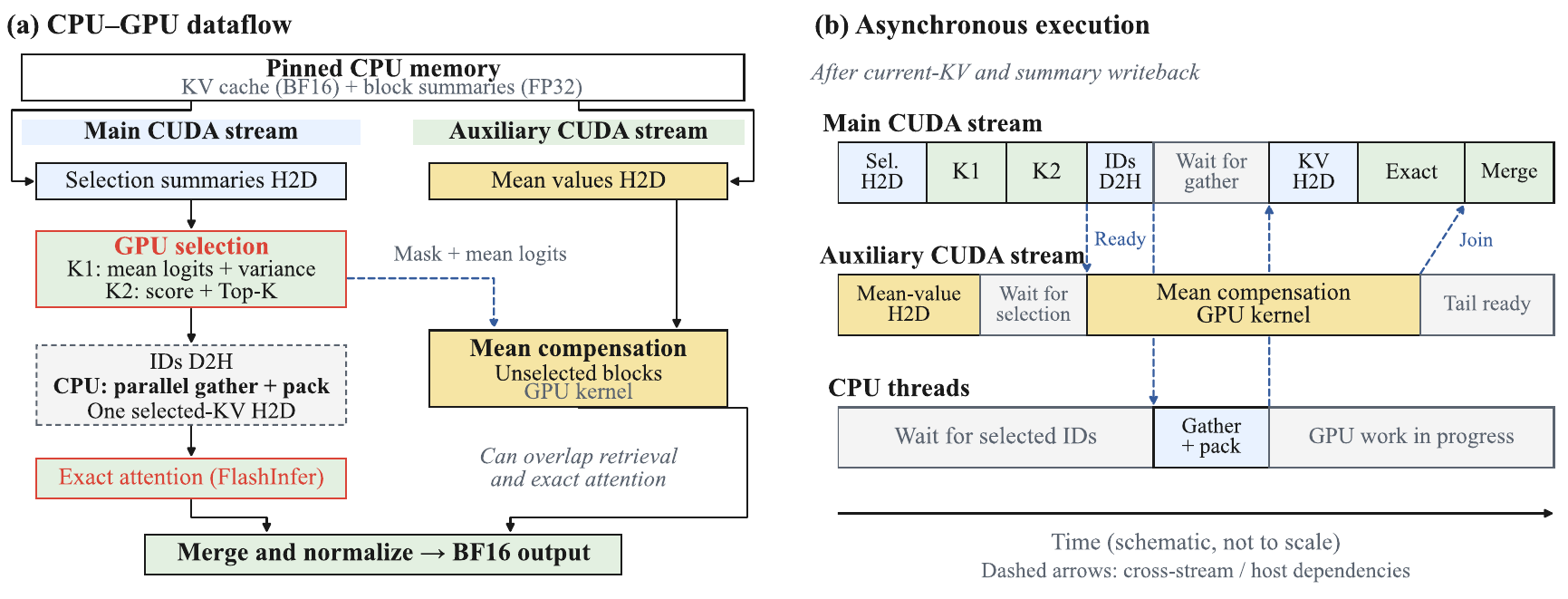}
  \caption{Implementation of CPU-offload CompKV. (a) Dataflow across the main stream, auxiliary stream, and CPU. (b) Asynchronous schedule: after mean values and selection are ready, Mean compensation runs on an auxiliary stream while the main path calculates the exact attention. The branches join at the final merge. Dashed arrows indicate dependencies; durations are schematic.}
  \label{fig:ours-offload-pipeline}
\end{figure*}

In practical LLM deployments, the KV cache may need to be stored outside GPU memory. Therefore, CompKV keeps the complete BF16 KV cache and FP32 block summaries in pinned CPU memory. Each step uses a fused GPU kernel to update the current block's statistics and pack the new K/V, followed by one device-to-host (D2H) transfer and CPU writeback. The summary buffer separates selection statistics from mean values. The main CUDA stream transfers the selection region and runs two kernels for block statistics and normalized Top-$K$ scoring, while an auxiliary stream independently transfers mean values. The selector reuses grouped variance projection in both factors of Equation~\ref{eq:compkv-score} to estimate both block mass and residual magnitude.

After selection, the CPU gathers the selected K/V into a contiguous pinned buffer for one host-to-device (H2D) transfer and FlashInfer exact attention. Meanwhile, the auxiliary stream computes Mean compensation once the mean values and selected-block mask are ready, overlapping this computation with CPU gathering, selected-KV transfer, and exact attention. The main stream waits for compensation immediately before merging both contributions under the shared normalization in Equation~\ref{eq:compkv-output}. Figure~\ref{fig:ours-offload-pipeline} shows the schedule; Appendix~\ref{app:implementation-details} details implementation and measurement.

\begin{table}[ht]
\centering
\caption{\textbf{RULER accuracy (\%).} Bold values mark the best sparse result per model and column, including ties. AVG annotations show drops ($\downarrow$) from Full in percentage points.}
\label{tab:ruler-main}
\scriptsize
\setlength{\tabcolsep}{1.25pt}
\renewcommand{\arraystretch}{1.22}
\newcommand{\rulerdelta}[3]{\makebox[3.1em][l]{\textcolor{#1}{\ensuremath{#2}\,{\fontsize{5}{6}\selectfont #3}}}}
\resizebox{\linewidth}{!}{%
\begin{tabular}{@{}l*{14}{r}@{}}
\toprule
\textbf{Method} & \textbf{SG1} & \textbf{SG2} & \textbf{SG3} & \textbf{MK1} & \textbf{MK2} & \textbf{MK3} & \textbf{MV} & \textbf{MQ} & \textbf{VT} & \textbf{CWE} & \textbf{FWE} & \textbf{QA1} & \textbf{QA2} & \multicolumn{1}{l}{\textbf{AVG}}\\

\midrule
\rowcolor{black!10}
\multicolumn{15}{c}{\strut\textit{Llama-3.1-8B-Instruct}}\\
Full & 100.0 & 100.0 & 100.0 & 100.0 & 99.8 & 99.2 & 94.9 & 99.1 & 99.1 & 9.9 & 93.3 & 81.4 & 54.4 & 87.0\phantom{\rulerdelta{red}{\downarrow}{00.0}}\\
\specialrule{\lightrulewidth}{0pt}{0pt}
Quest & \textbf{100.0} & \textbf{99.6} & 99.6 & \textbf{99.8} & 92.0 & 22.4 & 86.9 & 96.8 & 94.4 & 1.7 & 80.7 & 76.2 & 50.8 & 77.0\rulerdelta{red}{\downarrow}{10.0}\\
InfLLM & \textbf{100.0} & 95.4 & 78.2 & 91.0 & 92.0 & 38.0 & 64.0 & 83.0 & 90.6 & \textbf{2.7} & 84.4 & 80.0 & 51.4 & 73.1\rulerdelta{red}{\downarrow}{13.9}\\
Quest+RESA & \textbf{100.0} & 99.4 & 99.6 & \textbf{99.8} & 93.8 & 29.4 & \textbf{91.8} & \textbf{97.2} & 97.1 & 0.1 & 82.5 & 75.4 & 50.6 & 78.2\rulerdelta{red}{\downarrow}{8.8}\\
\textbf{CompKV} & \textbf{100.0} & \textbf{99.6} & \textbf{99.8} & 98.6 & \textbf{97.8} & \textbf{73.0} & 91.2 & 97.1 & \textbf{97.2} & 0.5 & \textbf{91.7} & \textbf{80.8} & \textbf{54.0} & \textbf{83.2}\rulerdelta{green!50!black}{\downarrow}{3.8}\\
\midrule
\rowcolor{black!10}
\multicolumn{15}{c}{\strut\textit{Qwen3-8B}}\\
Full & 100.0 & 100.0 & 100.0 & 98.6 & 97.4 & 98.8 & 96.2 & 97.5 & 100.0 & 82.7 & 92.3 & 71.8 & 54.0 & 91.5\phantom{\rulerdelta{red}{\downarrow}{00.0}}\\
\specialrule{\lightrulewidth}{0pt}{0pt}
Quest & \textbf{100.0} & \textbf{100.0} & 97.6 & 98.2 & 77.8 & 3.4 & 82.0 & 95.3 & 99.7 & 34.8 & 93.3 & 64.0 & 49.0 & 76.5\rulerdelta{red}{\downarrow}{15.0}\\
InfLLM & \textbf{100.0} & 94.0 & 76.0 & 94.0 & 89.6 & 42.8 & 81.9 & 85.9 & \textbf{99.8} & 20.0 & 88.4 & 70.6 & 52.6 & 76.6\rulerdelta{red}{\downarrow}{14.9}\\
Quest+RESA & \textbf{100.0} & \textbf{100.0} & 97.2 & 97.6 & 75.8 & 4.8 & 85.6 & 95.7 & 99.6 & 21.1 & 93.9 & 65.2 & 48.6 & 75.8\rulerdelta{red}{\downarrow}{15.7}\\
\textbf{CompKV} & \textbf{100.0} & \textbf{100.0} & \textbf{100.0} & \textbf{98.6} & \textbf{96.4} & \textbf{75.0} & \textbf{95.7} & \textbf{97.2} & \textbf{99.8} & \textbf{35.2} & \textbf{94.9} & \textbf{71.8} & \textbf{53.4} & \textbf{86.0}\rulerdelta{green!50!black}{\downarrow}{5.5}\\
\midrule
\rowcolor{black!10}
\multicolumn{15}{c}{\strut\textit{Qwen3-32B}}\\
Full & 100.0 & 98.4 & 100.0 & 99.8 & 99.8 & 100.0 & 99.4 & 100.0 & 99.7 & 87.5 & 93.3 & 80.0 & 60.4 & 93.7\phantom{\rulerdelta{red}{\downarrow}{00.0}}\\
\specialrule{\lightrulewidth}{0pt}{0pt}
Quest & \textbf{100.0} & \textbf{99.2} & 94.4 & \textbf{100.0} & 35.2 & 2.4 & 95.8 & \textbf{99.8} & 98.4 & 35.5 & 93.1 & 74.8 & 55.8 & 75.7\rulerdelta{red}{\downarrow}{18.0}\\
InfLLM & \textbf{100.0} & 95.8 & 82.4 & 93.4 & 91.6 & 68.0 & 85.3 & 90.1 & \textbf{99.4} & 34.9 & 93.3 & 81.0 & 57.0 & 82.5\rulerdelta{red}{\downarrow}{11.2}\\
Quest+RESA & 97.6 & 98.0 & 93.0 & \textbf{100.0} & 38.2 & 3.2 & 94.8 & 99.4 & 98.1 & 39.5 & 91.3 & 74.6 & 55.4 & 75.6\rulerdelta{red}{\downarrow}{18.1}\\
\textbf{CompKV} & \textbf{100.0} & \textbf{99.2} & \textbf{100.0} & 99.8 & \textbf{99.4} & \textbf{93.0} & \textbf{98.5} & 99.6 & 99.2 & \textbf{56.0} & \textbf{96.7} & \textbf{81.4} & \textbf{60.2} & \textbf{91.0}\rulerdelta{green!50!black}{\downarrow}{2.7}\\
\bottomrule
\end{tabular}}
\end{table}

\section{Experiments}
\subsection{Setup}
\label{sec:experiment-setup}

\paragraph{Models.} To comprehensively evaluate our proposed CompKV, we consider three models that are widely used in sparse attention studies: Llama-3.1-8B-Instruct~\citep{llamateam2024llama}, Qwen3-8B, and Qwen3-32B~\citep{qwen3technicalreport}. Qwen3-8B and Qwen3-32B are used to assess generalization across model scales, while Llama-3.1-8B-Instruct is included to further examine whether CompKV generalizes across different model families and architectural designs.

\paragraph{Methods.} To verify the effectiveness of CompKV, we compare it against Quest~\citep{tang2024quest}, InfLLM~\citep{xiao2024infllm}, and Quest+RESA with $\lambda=0.25$~\citep{yang2026resa}. Quest and InfLLM are state-of-the-art training-free, query-aware baselines for block selection. CompKV differs from these methods in two aspects: how blocks are selected for exact attention computation and how the unselected blocks are handled. RESA is a plug-in compensation method that can be integrated with different sparse-attention methods; following its original evaluation setup, we use Quest+RESA as the RESA baseline. As defined in Section~\ref{sec:compkv-score}, $r$ denotes the number of variance groups, and we set $r=4$ for all main CompKV results. Full attention is included as the dense reference.

\paragraph{Benchmarks and metrics.} RULER~\citep{hsieh2024ruler} provides controlled retrieval, tracing, and aggregation tasks to test whether CompKV preserves long-context capabilities under tight exact-read budgets. LongBench-Pro~\citep{chen2026longbenchpro} assesses whether these gains extend to more realistic tasks requiring either localized evidence retrieval or global context integration. We evaluate RULER at 32K and LongBench-Pro at context lengths up to 128K using their respective official scorers.

\paragraph{Additional configuration.} All evaluations are conducted on NVIDIA H100 80GB HBM3 GPUs with a batch size of one. For a consistent comparison, all four sparse methods partition consecutive tokens into blocks of size 16 and retain the first block as an attention sink together with the two most recent blocks as a local attention window. These mandatory blocks are always included in the selected set and count toward the same per-step exact-read budget. Appendix~\ref{app:eval-config} provides further details on the complete experimental setup and evaluation configurations.

\begin{table}[ht]
\centering
\caption{\textbf{LongBench-Pro scores.} Mod. and Extr. abbreviate Moderate and Extreme. Bold marks the best sparse result per model and column. AVG arrows show point drops from Full.}
\label{tab:longbench-main}
\normalsize
\setlength{\tabcolsep}{3.5pt}
\renewcommand{\arraystretch}{1.2}
\newcommand{\lbpdelta}[3]{\makebox[3.1em][l]{\textcolor{#1}{\ensuremath{#2}\,{\fontsize{6.4}{7.5}\selectfont #3}}}}
\begin{tabular*}{\linewidth}{@{\extracolsep{\fill}}l*{10}{r}@{}}
\toprule
\multirow{2}{*}{\textbf{Method}} & \multicolumn{4}{c}{\textbf{Difficulty}} & \multicolumn{5}{c}{\textbf{Context length}} & \multicolumn{1}{l}{\multirow{2}{*}{\textbf{AVG}}}\\
\cmidrule(lr){2-5}\cmidrule(lr){6-10}
 & \makebox[2.5em][r]{\textbf{Easy}} & \makebox[2.5em][r]{\textbf{Mod.}} & \makebox[2.5em][r]{\textbf{Hard}} & \makebox[2.5em][r]{\textbf{Extr.}} & \textbf{8k} & \textbf{16k} & \textbf{32k} & \textbf{64k} & \textbf{128k} & \\

\midrule
\rowcolor{black!10}
\multicolumn{11}{c}{\strut\textit{Llama-3.1-8B-Instruct}}\\
Full & 28.30 & 18.88 & 25.61 & 25.72 & 30.46 & 25.55 & 25.73 & 22.59 & 21.98 & 25.26\phantom{\lbpdelta{red}{\downarrow}{0.00}}\\
\specialrule{\lightrulewidth}{0pt}{0pt}
Quest & \textbf{28.30} & 17.71 & 22.66 & 24.75 & 28.09 & \textbf{25.54} & 24.47 & 21.55 & \textbf{21.75} & 24.28\lbpdelta{red}{\downarrow}{0.98}\\
InfLLM & 27.56 & 17.69 & 21.84 & 23.72 & 28.31 & 24.53 & 24.07 & 20.72 & 20.35 & 23.60\lbpdelta{red}{\downarrow}{1.66}\\
Quest+RESA & 27.51 & \textbf{18.09} & \textbf{24.38} & 24.54 & 27.95 & 24.34 & \textbf{25.94} & \textbf{21.73} & 21.54 & 24.30\lbpdelta{red}{\downarrow}{0.96}\\
\textbf{CompKV} & 27.91 & 18.05 & 23.94 & \textbf{24.83} & \textbf{28.63} & 25.36 & 25.30 & 21.50 & 21.39 & \textbf{24.44}\lbpdelta{green!50!black}{\downarrow}{0.82}\\
\midrule
\rowcolor{black!10}
\multicolumn{11}{c}{\strut\textit{Qwen3-8B}}\\
Full & 39.63 & 29.36 & 32.36 & 31.62 & 38.68 & 35.86 & 34.88 & 29.53 & 32.21 & 34.23\phantom{\lbpdelta{red}{\downarrow}{0.00}}\\
\specialrule{\lightrulewidth}{0pt}{0pt}
Quest & 34.54 & 27.14 & 28.68 & 28.89 & 33.83 & 29.54 & 30.50 & 27.84 & 31.19 & 30.58\lbpdelta{red}{\downarrow}{3.65}\\
InfLLM & 36.18 & 29.09 & 29.82 & 28.83 & 35.01 & 30.87 & 33.43 & 27.13 & 32.31 & 31.75\lbpdelta{red}{\downarrow}{2.48}\\
Quest+RESA & 34.54 & 27.82 & 28.63 & 28.46 & 33.90 & 29.96 & 31.37 & 27.25 & 30.50 & 30.60\lbpdelta{red}{\downarrow}{3.63}\\
\textbf{CompKV} & \textbf{37.17} & \textbf{30.09} & \textbf{30.68} & \textbf{29.89} & \textbf{35.74} & \textbf{32.82} & \textbf{33.77} & \textbf{28.17} & \textbf{33.19} & \textbf{32.74}\lbpdelta{green!50!black}{\downarrow}{1.49}\\
\midrule
\rowcolor{black!10}
\multicolumn{11}{c}{\strut\textit{Qwen3-32B}}\\
Full & 50.88 & 39.11 & 39.17 & 33.63 & 48.20 & 44.45 & 41.56 & 38.72 & 37.09 & 42.00\phantom{\lbpdelta{red}{\downarrow}{0.00}}\\
\specialrule{\lightrulewidth}{0pt}{0pt}
Quest & 44.93 & 33.93 & 35.76 & 31.43 & 41.67 & 35.15 & 39.00 & 36.22 & 36.06 & 37.62\lbpdelta{red}{\downarrow}{4.38}\\
InfLLM & 48.07 & \textbf{36.44} & 37.18 & 31.80 & 44.63 & 38.75 & \textbf{42.02} & 36.54 & 36.15 & 39.62\lbpdelta{red}{\downarrow}{2.38}\\
Quest+RESA & 43.61 & 31.93 & 35.11 & 29.69 & 41.87 & 35.24 & 38.35 & 35.07 & 30.28 & 36.16\lbpdelta{red}{\downarrow}{5.84}\\
\textbf{CompKV} & \textbf{49.56} & 35.23 & \textbf{38.62} & \textbf{32.36} & \textbf{45.49} & \textbf{39.52} & 41.09 & \textbf{38.28} & \textbf{37.15} & \textbf{40.30}\lbpdelta{green!50!black}{\downarrow}{1.70}\\
\bottomrule
\end{tabular*}
\end{table}

\subsection{Accuracy Results}
\label{sec:main-accuracy}

We evaluate how well CompKV preserves task quality by comparing overall scores, task-level behavior, and sensitivity to the KV budget. With $r=4$, CompKV achieves the highest AVG for all three models on RULER at a 32K length and a 512-token budget (Table~\ref{tab:ruler-main}), as well as on LongBench-Pro (Table~\ref{tab:longbench-main}). It also obtains the best or tied-best sparse scores in most task-level comparisons.

The gains are particularly pronounced on MK3, where the accuracy of CompKV is significantly higher than that of other baselines. This improvement suggests that compensation-aware selection helps preserve the information needed for difficult retrieval tasks. However, on Llama-3.1, Full attention scores only 9.9\% on CWE, and all evaluated sparse methods remain low, which means the aggregate improvements coexist with tasks that remain challenging even for the dense model.

The advantage persists as the exact-read budget varies. We further evaluate CompKV with token budgets ranging from 128 to 1024 for both 8B models on RULER. Results are shown in Appendix~\ref{app:budget-accuracy}. The curves show that several baselines approach Full as the budget increases, but CompKV remains the best across most budgets, tasks and models.

\subsection{Efficiency Evaluation}
\label{sec:efficiency-eval}

\begin{figure}[ht]
    \centering
    \includegraphics[width=0.98\textwidth]{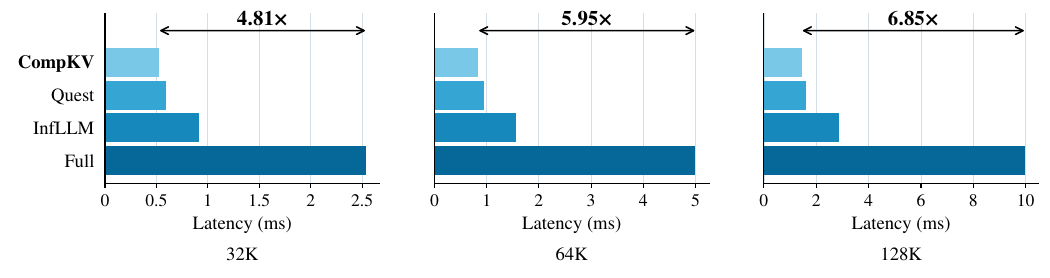}
    \caption{\textbf{CPU-offload attention latency.} Mean host wall time per single-layer attention step at a 512-token budget. Lengths denote the endpoints of the 256-step decode windows.}
    \label{fig:latency}
\end{figure}

To further evaluate the efficiency of our CompKV implementation, we follow the profiling setup of Quest~\citep{tang2024quest} and measure the complete CPU-offload attention step on a single NVIDIA H100 GPU with a batch size of one. We benchmark Full, Quest, InfLLM, and CompKV with $r\in\{4,128\}$ under the same evaluation setting. We do not include Quest+RESA in the latency evaluation because its original paper reports higher latency than Quest. The evaluation covers endpoint lengths of 32K, 64K, and 128K and selected-token budgets of 512, 1,024, and 2,048. CompKV ($r=4$) achieves the lowest mean latency in all nine settings, with up to $6.85\times$ speedups over Full. Figure~\ref{fig:latency} shows the 512-token budget; Appendix~\ref{app:latency-details} gives the full protocol and results.

\subsection{Ablation Studies}

\begin{wrapfigure}[28]{r}{0.45\columnwidth}
\vspace{-0.8\baselineskip}
\centering

\captionof{table}{\textbf{Factor ablations on RULER.} Values are 13-task AVG (\%) for Llama-3.1-8B-Instruct and Qwen3-8B with $r=4$. Column labels denote selected-token budgets.}
\label{tab:ruler-component-ablation}

\vspace{2pt}
\setlength{\tabcolsep}{2.5pt}

\begin{tabular*}{\linewidth}{
@{\extracolsep{\fill}}l*{4}{c}@{}
}
\toprule
\multirow{2}{*}{\textbf{Variant}}
& \multicolumn{2}{c}{\textbf{Llama}}
& \multicolumn{2}{c}{\textbf{Qwen}}\\
\cmidrule(lr){2-3}\cmidrule(lr){4-5}
& \textbf{256} & \textbf{512}
& \textbf{256} & \textbf{512}\\
\midrule
\textbf{CompKV}
& \textbf{78.6} & \textbf{83.2}
& \textbf{80.9} & \textbf{86.0}\\
$\widehat p^{(2)}$ only
& 75.6 & 81.8 & 79.2 & 83.5\\
$\widehat p^{(2)}$ + Mean
& 78.1 & 82.8 & 79.7 & 85.1\\
$\widehat p^{(1)}\widehat\sigma^2$ + Mean
& 76.1 & 81.2 & 77.5 & 83.8\\
\bottomrule
\end{tabular*}

\vspace{8pt}

\includegraphics[
  width=\linewidth
]{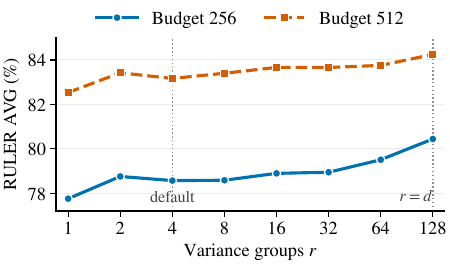}

\vspace{-4pt}
\captionof{figure}{\textbf{Variance-group granularity on RULER.}
Curves show the 13-task AVG for Llama-3.1-8B-Instruct at
selected-token budgets of 256 and 512.}
\label{fig:variance-group-granularity}

\vspace{-0.5\baselineskip}
\end{wrapfigure}

\paragraph{Ablation objectives.}
We examine Mean compensation, the outer variance factor, and the second-order mass correction to determine how tail reconstruction and residual-aware scoring contribute to the final task quality. We also vary the number of variance groups $r$ to assess how compact block statistics affect accuracy and metadata cost.

\paragraph{Factor ablations.}
\label{sec:component-ablations}
Table~\ref{tab:ruler-component-ablation} compares four configurations with $r=4$. Here, $\widehat p^{(2)}$ includes the variance correction in Equation~\ref{eq:compkv-mass-estimate}, while $\widehat p^{(1)}$ uses mean-key logits alone. Adding Mean compensation improves AVG by 0.48--2.47 points, while the outer variance factor provides a further 0.35--1.19 points. The second-order mass correction contributes an additional 1.96--3.39 points over its first-order counterpart. Overall, the results indicate that compensation, residual-aware selection, and second-order mass estimation all contribute consistently, with CompKV achieving the highest AVG in all four settings.

\paragraph{Variance-group granularity.}
\label{sec:variance-group-granularity}
Figure~\ref{fig:variance-group-granularity} shows that accuracy generally improves with finer variance groups, with $r=128$ achieving the highest AVG at both budgets. Our default $r=4$ trails this endpoint by 1.87 and 1.08 points at budgets 256 and 512, respectively, while requiring only $1/32$ of its variance metadata. This trade-off motivates the use of compact grouped statistics in the main experiments.

\section{Conclusion}

In this paper, we identify the mismatch between KV block selection and downstream compensation and introduce CompKV, a compensation-aware method that selects blocks according to their estimated post-compensation residual. Our analysis shows that, under Mean compensation, this residual is jointly governed by block attention mass and within-block logit variation, which CompKV estimates from compact block statistics. Across three models and two long-context benchmarks, CompKV achieves the highest average scores among the evaluated sparse methods, while its asynchronous implementation delivers up to $6.85\times$ self-attention speedups over full attention.


\bibliography{iclr2027_conference}
\bibliographystyle{iclr2027_conference}

\appendix
\section{Detailed CompKV Derivations}
\label{app:compkv-derivations}

\subsection{KL Identity for Mean Compensation}
\label{app:compkv-kl-proofs}

Mean compensation replaces the exact weight of each unselected block with its mean-based weight. Consequently, the selection loss admits both a log-partition form and a form that isolates the unreconstructed weight of each omitted block (Equation~\ref{eq:mean-residual-objective}):
\begin{equation*}
\begin{aligned}
\mathcal L(\mathcal S)
&=
\sum_{g=1}^{G}
\log
\frac{\displaystyle\sum_{b\in\mathcal B}Z_{g,b}}
{\displaystyle\sum_{b\in\mathcal S}Z_{g,b}
+\sum_{b\notin\mathcal S}|b|e^{\bar z_{g,b}}}\\
&=
-\sum_{g=1}^{G}
\log\left[
1-\sum_{b\notin\mathcal S}p_{g,b}
\left(1-\frac{|b|e^{\bar z_{g,b}}}{Z_{g,b}}\right)
\right].
\end{aligned}
\end{equation*}

To establish this identity, fix a query head $g$ and a selected block set $\mathcal S$. Write the full and compensated normalization constants as
\begin{equation}
D_g=\sum_{b\in\mathcal B}Z_{g,b},\qquad
\widehat D_g(\mathcal S)=
\sum_{b\in\mathcal S}Z_{g,b}
+\sum_{b\notin\mathcal S}|b|e^{\bar z_{g,b}}.
\end{equation}
The token probabilities are $P_g(j)=e^{z_{g,j}}/D_g$ and $\widehat P_g(j;\mathcal S)=e^{\hat z_{g,j}}/\widehat D_g(\mathcal S)$. Substituting them into the KL divergence gives
\begin{equation}
\label{eq:app-mean-log-ratio}
\begin{aligned}
D_{\mathrm{KL}}\!\left(\widehat P_g(\mathcal S)\Vert P_g\right)
&=
\sum_{b\in\mathcal B}\sum_{j\in b}
\widehat P_g(j;\mathcal S)
\left[
\hat z_{g,j}-z_{g,j}
+\log\frac{D_g}{\widehat D_g(\mathcal S)}
\right]\\
&=
\log\frac{D_g}{\widehat D_g(\mathcal S)}
+\sum_{b\notin\mathcal S}\sum_{j\in b}
\widehat P_g(j;\mathcal S)
\left(\bar z_{g,b}-z_{g,j}\right).
\end{aligned}
\end{equation}
The second equality follows from $\hat z_{g,j}=z_{g,j}$ in selected blocks and normalized probabilities.

Within each unselected block, every token has the same compensated probability. The remaining term therefore vanishes:
\begin{equation}
\label{eq:app-mean-cancellation}
\sum_{b\notin\mathcal S}\sum_{j\in b}
\widehat P_g(j;\mathcal S)
\left(\bar z_{g,b}-z_{g,j}\right)\\
=
\frac{1}{\widehat D_g(\mathcal S)}
\sum_{b\notin\mathcal S}e^{\bar z_{g,b}}
\sum_{j\in b}\left(\bar z_{g,b}-z_{g,j}\right)
=0,
\end{equation}
because $\bar z_{g,b}$ is the arithmetic mean of the original logits in block $b$. Summing the resulting log-partition ratio over the $G$ query heads yields
\begin{equation}
\mathcal L(\mathcal S)
=
\sum_{g=1}^{G}\log\frac{D_g}{\widehat D_g(\mathcal S)}
=
\sum_{g=1}^{G}
\log
\frac{\displaystyle\sum_{b\in\mathcal B}Z_{g,b}}
{\displaystyle
\sum_{b\in\mathcal S}Z_{g,b}
+\sum_{b\notin\mathcal S}|b|e^{\bar z_{g,b}}
}.
\end{equation}

To obtain the residual form, write the compensated normalization constant as the full weight minus the unreconstructed weight:
\begin{equation}
\widehat D_g(\mathcal S)
=
D_g-\sum_{b\notin\mathcal S}
\left(Z_{g,b}-|b|e^{\bar z_{g,b}}\right).
\end{equation}
Dividing by $D_g$ and using $p_{g,b}=Z_{g,b}/D_g$ gives
\begin{equation}
\frac{\widehat D_g(\mathcal S)}{D_g}
=
1-\sum_{b\notin\mathcal S}
p_{g,b}
\left(1-\frac{|b|e^{\bar z_{g,b}}}{Z_{g,b}}\right).
\end{equation}
Substitution into the log-partition ratio establishes
\begin{equation}
\mathcal L(\mathcal S)
=
-\sum_{g=1}^{G}
\log\left[
1-\sum_{b\notin\mathcal S}
p_{g,b}
\left(1-\frac{|b|e^{\bar z_{g,b}}}{Z_{g,b}}\right)
\right],
\end{equation}
which is the second equality in Equation~\ref{eq:mean-residual-objective}.

\subsection{Expansion of the Blockwise Compensation Residual}
\label{app:compkv-block-residual}

The residual form above shows that an omitted block contributes to the loss only through the weight that Mean compensation fails to recover. In the small-variance regime, this unreconstructed fraction has the expansion in Equation~\ref{eq:mean-residual-expansion}:
\begin{equation*}
p_{g,b}\left(1-\frac{|b|e^{\bar z_{g,b}}}{Z_{g,b}}\right)
=
p_{g,b}\left(\frac12\sigma_{g,b}^2+O(\sigma_{g,b}^3)\right).
\end{equation*}

To establish this expansion, we hold the maximum block size fixed and let $\max_{g,b}\sigma_{g,b}\to0$, where $\sigma_{g,b}$ is the standard deviation of the logits within block $b$ for query head $g$. We start from the block's exact weight relative to its mean-based weight.

Averaging the second-order Taylor expansion of the exponential over a block gives
\begin{equation}
\label{eq:mean-partition-expansion}
\frac{Z_{g,b}}{|b|e^{\bar z_{g,b}}}=\frac1{|b|}\sum_{j\in b}e^{z_{g,j}-\bar z_{g,b}}=1+\frac12\sigma_{g,b}^2+O(\sigma_{g,b}^3).
\end{equation}
The linear term vanishes because $\sum_{j\in b}(z_{g,j}-\bar z_{g,b})=0$, and the quadratic term averages to $\sigma_{g,b}^2/2$. The fixed maximum block size makes the averaged third-order remainder $O(\sigma_{g,b}^3)$.

The deviation of this ratio from one is $O(\sigma_{g,b}^2)$, so taking its reciprocal gives
\begin{equation}
\frac{|b|e^{\bar z_{g,b}}}{Z_{g,b}}=\frac1{1+\frac12\sigma_{g,b}^2+O(\sigma_{g,b}^3)}=1-\frac12\sigma_{g,b}^2+O(\sigma_{g,b}^3).
\end{equation}
The reciprocal expansion introduces a quadratic remainder of order $O(\sigma_{g,b}^4)$, which combines with the existing cubic remainder. Subtracting the reciprocal from one and multiplying by $p_{g,b}$ yields
\begin{equation}
p_{g,b}\left(1-\frac{|b|e^{\bar z_{g,b}}}{Z_{g,b}}\right)=p_{g,b}(\frac12\sigma_{g,b}^2+O(\sigma_{g,b}^3)).
\end{equation}

Taking the logarithm of Equation~\ref{eq:mean-partition-expansion} also gives the log-partition expansion used for attention-mass estimation:
\begin{equation}
\label{eq:compkv-log-partition}
\log Z_{g,b}=\log|b|+\bar z_{g,b}+\frac12\sigma_{g,b}^2+O(\sigma_{g,b}^3).
\end{equation}

\subsection{Second-Order Expansion of the Selection Loss}
\label{app:compkv-taylor}

The preceding subsection expands the residual of a single omitted block. We now combine these residuals within each query head and account for the outer logarithm in the exact selection loss. This gives the second-order approximation in Equation~\ref{eq:mean-residual-approximation}:
\begin{equation*}
\mathcal L(\mathcal S)
=
\frac12\sum_{g=1}^{G}\sum_{b\notin\mathcal S}
p_{g,b}\left(\sigma_{g,b}^2+O\!\left(\sigma_{g,b}^3\right)\right).
\end{equation*}

To derive this expression, hold the maximum block size fixed and set $\varepsilon=\max_{g,b}\sigma_{g,b}\to0$. Define the total normalized
residual for head $g$ as
\begin{equation}
R_g(\mathcal S)
=
\sum_{b\notin\mathcal S}
p_{g,b}\left(
1-\frac{|b|e^{\bar z_{g,b}}}{Z_{g,b}}
\right).
\end{equation}
Summing the blockwise expansion from
Appendix~\ref{app:compkv-block-residual} gives
\begin{equation}
R_g(\mathcal S)
=
\frac12\sum_{b\notin\mathcal S}p_{g,b}\sigma_{g,b}^2
+
O\!\left(
\sum_{b\notin\mathcal S}p_{g,b}\sigma_{g,b}^3
\right).
\end{equation}
Since
$\sum_{b\notin\mathcal S}p_{g,b}\sigma_{g,b}^3
\le
\varepsilon\sum_{b\notin\mathcal S}p_{g,b}\sigma_{g,b}^2$,
we have $R_g=O(\sum_{b\notin\mathcal S}p_{g,b}\sigma_{g,b}^2)$. Moreover, $\sum_{b\notin\mathcal S}p_{g,b}\le1$ implies $R_g\to0$. Expanding the outer logarithm therefore yields
\begin{equation}
\begin{aligned}
-\log(1-R_g)
={}&
\frac12\sum_{b\notin\mathcal S}p_{g,b}\sigma_{g,b}^2\\
&+
O\!\left(
\sum_{b\notin\mathcal S}p_{g,b}\sigma_{g,b}^3
+
\left[
\sum_{b\notin\mathcal S}p_{g,b}\sigma_{g,b}^2
\right]^2
\right).
\end{aligned}
\end{equation}

The quadratic remainder can be absorbed into the cubic term. By the Cauchy--Schwarz inequality,
\begin{equation}
\begin{aligned}
\left[
\sum_{b\notin\mathcal S}p_{g,b}\sigma_{g,b}^2
\right]^2
&\le
\left(\sum_{b\notin\mathcal S}p_{g,b}\right)
\left(\sum_{b\notin\mathcal S}p_{g,b}\sigma_{g,b}^4\right)\\
&\le
\varepsilon
\sum_{b\notin\mathcal S}p_{g,b}\sigma_{g,b}^3.
\end{aligned}
\end{equation}
Substituting this bound and summing over the query heads gives
\begin{equation}
\begin{aligned}
\mathcal L(\mathcal S)
&=
\frac12\sum_{g=1}^{G}\sum_{b\notin\mathcal S}
p_{g,b}\sigma_{g,b}^2
+
O\!\left(
\sum_{g=1}^{G}\sum_{b\notin\mathcal S}
p_{g,b}\sigma_{g,b}^3
\right)\\
&=
\frac12\sum_{g=1}^{G}\sum_{b\notin\mathcal S}
p_{g,b}\left(
\sigma_{g,b}^2+O(\sigma_{g,b}^3)
\right),
\end{aligned}
\end{equation}
which establishes Equation~\ref{eq:mean-residual-approximation}. The remainder bounds hold uniformly over the selected sets throughout the stated small-variance regime with a fixed maximum block size.

The leading term depends on $\mathcal S$ only through the unselected blocks. Because its sum over all blocks is independent of $\mathcal S$, minimizing the unselected contribution under a budget of $K$ exact block reads is equivalent to
\begin{equation*}
\max_{\mathcal S\subseteq\mathcal B,\,|\mathcal S|=K}
\sum_{b\in\mathcal S}\sum_{g=1}^{G}p_{g,b}\sigma_{g,b}^2.
\end{equation*}
Thus, the second-order objective selects the $K$ blocks with the largest
$\sum_{g=1}^{G}p_{g,b}\sigma_{g,b}^2$, as stated in
Equation~\ref{eq:mean-second-order-selection}.

\subsection{Grouped Variance and Attention-Mass Estimation}
\label{app:compkv-grouped}

The oracle selection rule above requires each block's exact attention mass and logit variance, which would require reading its keys. CompKV instead estimates the variance from grouped key statistics. The first quantity to derive is Equation~\ref{eq:grouped-variance-metadata}:
\begin{equation*}
\widehat\sigma_{g,b}^2
=
\frac1d\sum_{t=0}^{r-1}
\left(\sum_{i\in\mathcal G_t}q_{g,i}^2\right)(\bar{\mathbf D}_b)_t,
\qquad
(\bar{\mathbf D}_b)_t
=
\frac1{|b|\,|\mathcal G_t|}
\sum_{j\in b}\sum_{i\in\mathcal G_t}
(k_{j,i}-\bar k_{b,i})^2.
\end{equation*}

To obtain this estimate, let $\boldsymbol\Sigma_{k,b}\in\mathbb R^{d\times d}$
denote the key covariance within block $b$. Its definition and projection
along the query give
 
\begin{equation}
\label{eq:app-key-covariance}
\boldsymbol\Sigma_{k,b}
=
\frac1{|b|}\sum_{j\in b}
(\vk_j-\bar{\vk}_b)^\top(\vk_j-\bar{\vk}_b),
\qquad
\sigma_{g,b}^2
=
\frac1d\vq_g\boldsymbol\Sigma_{k,b}\vq_g^\top.
\end{equation}
The projection follows from $z_{g,j}-\bar z_{g,b}=\vq_g(\vk_j-\bar{\vk}_b)^\top/\sqrt d$.

For the settings where $r$ divides $d/2$, we divide the ordered RoPE frequency pairs into $r$ equal, consecutive groups, keeping both coordinates of each pair together. Each group $\mathcal G_t$ contains $d/r$ coordinates. At $r=d$, each group contains one coordinate. The statistic stored for each group is the average of its diagonal covariance entries:
\begin{equation}
(\bar{\mathbf D}_b)_t
=
\frac1{|\mathcal G_t|}
\sum_{i\in\mathcal G_t}(\boldsymbol\Sigma_{k,b})_{ii}
=
\frac1{|b|\,|\mathcal G_t|}
\sum_{j\in b}\sum_{i\in\mathcal G_t}
(k_{j,i}-\bar k_{b,i})^2.
\end{equation}

We approximate $\boldsymbol\Sigma_{k,b}$ by a diagonal matrix, assigning $(\bar{\mathbf D}_b)_t$ to every diagonal entry whose coordinate belongs to $\mathcal G_t$. Projecting this approximation along the query yields
\begin{equation}
\widehat\sigma_{g,b}^2
=
\frac1d\sum_{t=0}^{r-1}\sum_{i\in\mathcal G_t}
q_{g,i}^2(\bar{\mathbf D}_b)_t
=
\frac1d\sum_{t=0}^{r-1}
\left(\sum_{i\in\mathcal G_t}q_{g,i}^2\right)
(\bar{\mathbf D}_b)_t.
\end{equation}
Together, these expressions recover both parts of Equation~\ref{eq:grouped-variance-metadata}. Each group's average key variance is weighted by the sum of squared query coordinates in that group, with attention scaling $1/d$.

The same variance estimate also determines the attention-mass estimate in Equation~\ref{eq:compkv-mass-estimate}. Exponentiating the log-partition expansion in Equation~\ref{eq:compkv-log-partition} gives
\begin{equation}
Z_{g,b}
=
|b|\exp\!\left(
\bar z_{g,b}+\frac12\sigma_{g,b}^2+O(\sigma_{g,b}^3)
\right).
\end{equation}
Retaining the second-order terms and replacing $\sigma_{g,b}^2$ with $\widehat\sigma_{g,b}^2$ gives the estimated block weight $|b|\exp(\bar z_{g,b}+\widehat\sigma_{g,b}^2/2)$. Since the full attention mass is $p_{g,b}=Z_{g,b}/\sum_{c\in\mathcal B}Z_{g,c}$, normalizing these estimated weights over all available blocks yields
\begin{equation}
\widehat p_{g,b}
=
\frac{|b|\exp\!\left(\bar z_{g,b}+\frac12\widehat\sigma_{g,b}^2\right)}
{\displaystyle\sum_{c\in\mathcal B}|c|
\exp\!\left(\bar z_{g,c}+\frac12\widehat\sigma_{g,c}^2\right)}.
\end{equation}
This recovers Equation~\ref{eq:compkv-mass-estimate}, with normalization performed separately for each query head.

\subsection{Comparison with Uncompensated Sparse Attention}
\label{app:compkv-drop}

For comparison, Drop removes unselected blocks and renormalizes over the tokens in a nonempty selected set $\mathcal S$, giving $P_g^{\mathrm{Drop}}(\mathcal S)$. Unselected positions in the original token space receive zero probability. The ratio between the approximate and full probabilities is constant on selected positions:
\begin{equation}
\label{eq:app-drop-kl}
D_{\mathrm{KL}}\!\left(P_g^{\mathrm{Drop}}(\mathcal S)\Vert P_g\right)=\log\frac{\displaystyle\sum_{c\in\mathcal B}Z_{g,c}}{\displaystyle\sum_{b\in\mathcal S}Z_{g,b}}=-\log\!\left(\sum_{b\in\mathcal S}p_{g,b}\right).
\end{equation}

This identity agrees with the KL analysis of Top-$k$ sparse attention~\citep{tzachristas2026mathematical}. For one head and a fixed block budget, retaining the blocks with the largest attention mass exactly minimizes this KL divergence. With a shared selected set across heads, the objective sums the headwise logarithmic terms. When the omitted attention mass is small for each head, its first-order additive score is $\sum_g p_{g,b}$.

For the same nonempty selected set, adding positive mean-compensation weights increases the normalization constant. The inequality $|b|e^{\bar z_{g,b}}\le Z_{g,b}$ also ensures that this constant does not exceed that of full attention. Consequently,
\begin{equation}
\label{eq:app-mean-drop-ordering}
0\le D_{\mathrm{KL}}\!\left(\widehat P_g(\mathcal S)\Vert P_g\right)\le D_{\mathrm{KL}}\!\left(P_g^{\mathrm{Drop}}(\mathcal S)\Vert P_g\right).
\end{equation}

The second inequality is strict whenever an unselected block exists. The first is also strict if at least one unselected block has nonconstant logits. This comparison holds for a fixed selected set; the two operators can induce different optimal block sets under the same exact-read budget.

\section{Attention Fidelity of Residual-Aware Selection}
\label{app:residual-selection-diagnostics}

\subsection{Exact Residual versus Attention Mass}
\label{app:exact-residual-selection}
We compare exact mass-based and residual-based selection on Llama-3.1-8B-Instruct across all 13 RULER tasks, randomly sampling 64 prompts per task with seed 42 (832 total). We collect full-attention states under greedy decoding with batch size one at some layers. Both selectors use identical Q/K/V states, Mean compensation, and a 512-token budget with 16-token blocks; the first block and two most recent blocks are mandatory and included in the budget.

The selectors rank blocks by $\sum_g p_{g,b}$ or $\sum_g\rho_{g,b}$, sharing each selected set across heads, where
\begin{equation}
\label{eq:diagnostic-block-residual}
\rho_{g,b}
=
p_{g,b}\left(1-\frac{|b|e^{\bar z_{g,b}}}{Z_{g,b}}\right)
\end{equation}
is the exact compensation residual in Equation~\ref{eq:mean-residual-objective}. For each head $h$, we measure
\begin{equation}
\label{eq:residual-diagnostic-errors}
E_{\mathrm{KL},h}
=
D_{\mathrm{KL}}(\widehat P_h\Vert P_h),
\qquad
E_{\mathrm{out},h}
=
\frac{\|\widehat P_hV_h-P_hV_h\|_2}{\|P_hV_h\|_2}.
\end{equation}
KL compares post-softmax probabilities; relative $L_2$ compares outputs after multiplication by $V_h$. We plot $\Delta E_h=E_h^{(\mathrm{mass})}-E_h^{(\mathrm{res})}$, so \textbf{positive values favor residual-based selection}. Each point averages positions within prompts, prompts within tasks, and then all 13 tasks equally.

\begin{figure}[ht]
  \centering
  \includegraphics[width=\linewidth]{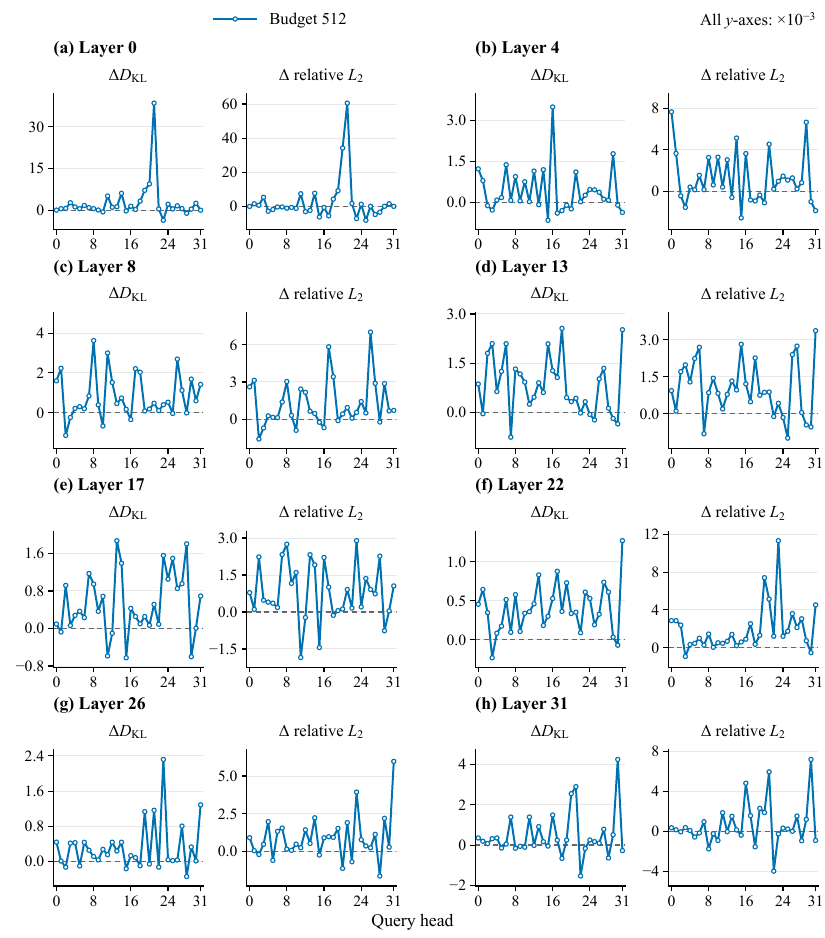}
  \caption{\textbf{Per-head error reduction from residual-based selection.} Mass-minus-residual differences in attention KL (left) and relative output $L_2$ (right), using the same Mean compensator.}
  \label{fig:compkv-residual-vs-mass-heads}
\end{figure}

Averaged over the eight layers and 32 query heads, residual-based selection reduces KL by 0.660\% and relative $L_2$ by 0.582\%. Both metrics improve in every layer-level mean, with improvements in most layer--head pairs. Thus, prioritizing the contribution left unreconstructed by Mean compensation lowers average attention and output error relative to mass-based selection.

Although the local gains are modest, attention-output errors propagate through subsequent layers and decoding steps and may be amplified, while reducing them may help limit accumulated deviations.

\subsection{Fidelity of Practical CompKV Scores}
\label{app:practical-score-fidelity}

On the same states and budget, we compare CompKV ($r\in\{4,128\}$) with estimated-mass selection, $\sum_g\widehat p_{g,b}$, using the $r=4$ mass estimate in Equation~\ref{eq:compkv-mass-estimate}. This baseline retains the second-order mass correction and removes the external variance factor in Equation~\ref{eq:compkv-score}.

For each KV group, let $T_b=\sum_g\rho_{g,b}$, with $\rho_{g,b}$ defined in Equation~\ref{eq:diagnostic-block-residual}. Spearman correlation compares scores with $T_b$ over non-mandatory blocks. Let $\mathcal S_m$ and $\mathcal S_\rho$ contain the 29 non-mandatory blocks selected by method $m$ and exact residual ranking. Residual capture is
\begin{equation}
\operatorname{Capture}(m)=
\frac{\sum_{b\in\mathcal S_m}T_b}
     {\sum_{b\in\mathcal S_\rho}T_b}.
\end{equation}
We use the averaging in Appendix~\ref{app:exact-residual-selection}. Layer curves additionally average KV heads, and overall results also average layers.

\begin{figure}[ht]
  \centering
  \includegraphics[width=\linewidth]{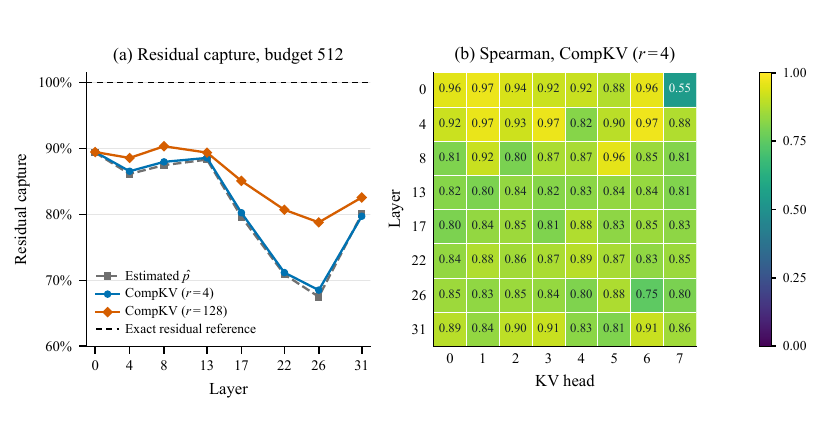}
  \caption{\textbf{Fidelity of practical CompKV scores.} (a) residual capture at budget 512, normalized by exact residual selection. (b) CompKV ($r=4$) Spearman correlations by layer and KV head. Mandatory blocks are excluded from both metrics.}
  \label{fig:practical-score-fidelity}
\end{figure}

Figure~\ref{fig:practical-score-fidelity} reports both metrics. Overall Spearman correlations are 0.8511, 0.8604, and 0.8782 for estimated mass, CompKV ($r=4$), and CompKV ($r=128$), respectively, with capture rates of 81.17\%, 81.53\%, and 85.61\%. CompKV ($r=4$) improves residual capture over estimated-mass selection in seven of eight layer means. Spearman correlations were saved as online scalars.

These results show that highly compressed block summaries preserve much of the residual-selection signal. With only four grouped-variance statistics per block, CompKV achieves strong ranking agreement and captures over 81\% of the residual removable by the exact reference. Its modest improvements over estimated-mass selection use the same summary statistics, showing the benefit of compensation-aware scoring within a fixed metadata budget. Although a gap to exact selection remains, this provides a practical balance between summary compactness and selection fidelity.

\section{Evaluation Configuration}
\label{app:eval-config}

\subsection{Model Configurations}

\begin{itemize}
\item \textbf{Llama-3.1-8B-Instruct} has 32 Transformer layers, 32 query heads, 8 key--value (KV) heads, a head dimension of 128, and a native context length of 131,072 tokens.
\item \textbf{Qwen3-8B} has 36 Transformer layers, 32 query heads, 8 KV heads, a head dimension of 128, and a native context length of 32,768 tokens.
\item \textbf{Qwen3-32B} has 64 Transformer layers, 64 query heads, 8 KV heads, a head dimension of 128, and a native context length of 32,768 tokens.
\end{itemize}

\subsection{Evaluation Benchmarks}

\paragraph{RULER.} RULER~\citep{hsieh2024ruler} is a synthetic long-context benchmark organized around retrieval, multi-hop tracing, aggregation, and question answering. We evaluate all 13 tasks in its 32K configuration, using 500 samples per task and 6,500 samples in total. Retrieval is measured by eight Needle-in-a-Haystack tasks: \texttt{niah\_single\_1/2/3}, \texttt{niah\_multikey\_1/2/3}, \texttt{niah\_multivalue}, and \texttt{niah\_multiquery}. \texttt{vt} evaluates variable tracking, \texttt{cwe} and \texttt{fwe} evaluate common- and frequent-word aggregation, and \texttt{qa\_1} and \texttt{qa\_2} evaluate question answering over the SQuAD and HotpotQA documents provided for these two tasks, respectively.

\paragraph{LongBench-Pro.} LongBench-Pro~\citep{chen2026longbenchpro} is constructed from natural long documents and contains 1,500 bilingual samples across 11 primary tasks and 25 secondary tasks. Its primary tasks are Retrieval \& Ranking, Sequencing \& Structure Reconstruction, Evidence-Grounded QA, Summarization \& Synthesis, Attribution \& Citation Alignment, Aggregation \& Clustering, Consistency \& Compliance Checking, Structured \& Numeric Reasoning, Version \& Code Diff Analysis, Rule Induction \& In-Context Learning, and Dialogue Memory \& Long-Horizon Tracking. We use the 625 English samples in the 8k, 16k, 32k, 64k, and 128k buckets, with 125 samples in each bucket. The selected samples cover both Full (global integration) and Partial (localized retrieval) context requirements and retain the Easy, Moderate, Hard, and Extreme difficulty labels.

\subsection{Evaluation Details}

\paragraph{InfLLM configuration.} We adapt InfLLM to 16-token blocks and grouped-query attention. Each query head accumulates causal attention probabilities and represents each block by the FP32 mean of its four highest-weight post-RoPE keys. Representatives are frozen when blocks leave the recent window, before selection. Query--representative scores are summed within each KV group to obtain a shared selected set. We retain the model's positional encoding, including YaRN for Qwen3 on LongBench-Pro, to compare KV selection and compensation under the positional semantics of the same Full attention reference. InfLLM's distant-position remapping changes both retrieval scores and attention logits. Retaining the configured positions keeps this factor consistent across methods and between dense prefill and decoding, allowing us to evaluate its representative-based selection within our common decode setting. Original Q/K/V states are BF16; representatives, accumulated weights, scores, and attention accumulation use FP32.

\paragraph{Prompt formatting.} Llama-3.1-8B-Instruct uses the official raw RULER prompt, in which the input is followed directly by the answer prefix. For Qwen3-8B and Qwen3-32B, the RULER input is placed in a user message and the answer prefix in the final assistant message, with generation continuing from that prefix by setting \texttt{continue\_final\_message=true}. For LongBench-Pro, we concatenate each context with the official \texttt{question\_nonthinking} field using four newline characters and then render the model-native chat template. Llama uses its instruct template, while both Qwen3 models use their assistant-prefill template with the generation prompt enabled.

\paragraph{No-thinking policy.} We evaluate both Qwen3 checkpoints in their supported non-thinking mode on RULER and LongBench-Pro, setting \texttt{enable\_thinking=False} in the chat template and \texttt{add\_special\_tokens=False} during tokenization. This establishes a direct-answer setting for comparing KV-selection methods under fixed output budgets. In preliminary runs, some methods still generated thinking markers despite this configuration. We therefore additionally suppress the \texttt{<think>} and \texttt{</think>} token IDs, 151667 and 151668, at every decoding step before token selection. This policy applies identically to Full attention, all sparse baselines, and CompKV.

\paragraph{Context handling.} RULER prompts are left-truncated only after reserving the task-specific completion length. On LongBench-Pro, both Qwen3 models use YaRN with a factor of 4 and an original maximum position of 32,768, giving an effective context length of 131,072 tokens, while Llama uses its native context window. If a rendered prompt plus 1,024 completion tokens exceeds the model context, we retain equal-length prefix and suffix segments through middle truncation.

\paragraph{Sampling and stopping.} RULER uses deterministic greedy decoding without sampling filters. Its official generation limits are 128 tokens for the retrieval tasks, 30 for \texttt{vt}, 120 for \texttt{cwe}, 50 for \texttt{fwe}, and 32 for each question-answering task. LongBench-Pro uses stochastic decoding: both Qwen3 models set $(\texttt{temperature},\texttt{top\_p},\texttt{top\_k},\texttt{min\_p})=(0.7,0.8,20,0)$, while Llama uses $(0.6,0.9,0,\texttt{None})$. All three models use \texttt{max\_new\_tokens=1024}, one beam, repetition penalty 1.0, and EOS-only stopping. RULER uses seed 42 for its single deterministic decode. Each LongBench-Pro case is decoded for three rounds with seeds 42, 43, and 44.

\paragraph{LongBench-Pro budgets.} The two token budgets are $(256,512)$ for 8k and 16k, $(512,1024)$ for 32k, $(1024,2048)$ for 64k, and $(2048,4096)$ for 128k. For all sparse methods, these values cap the exact KV tokens read per decoding step, including the sink and recent blocks.

\paragraph{Scoring metrics.} The official RULER scorer applies all-reference substring matching to the retrieval, tracing, and aggregation tasks and any-reference substring matching to the two question-answering tasks. The LongBench-Pro scorer selects the metric associated with each secondary task, including NDCG, Pairwise Accuracy, Accuracy, Summary, F1, and SubEM. Summary uses Qwen3-Embedding-8B as specified by the benchmark.

\paragraph{Score aggregation.} For RULER, we first average over the 500 samples within each task and then report the unweighted mean of the 13 task scores. For LongBench-Pro, we average the scores from the three rounds for each sample and budget tier, average the two tiers with equal weight, and then take the arithmetic mean over all 625 samples to obtain the AVG score reported in Table~\ref{tab:longbench-main}. Difficulty and length-bucket scores apply the same aggregation to their corresponding sample subsets.

\section{Accuracy across KV Budgets}
\label{app:budget-accuracy}

\begin{figure}[ht]
\centering
\includegraphics[width=\linewidth]{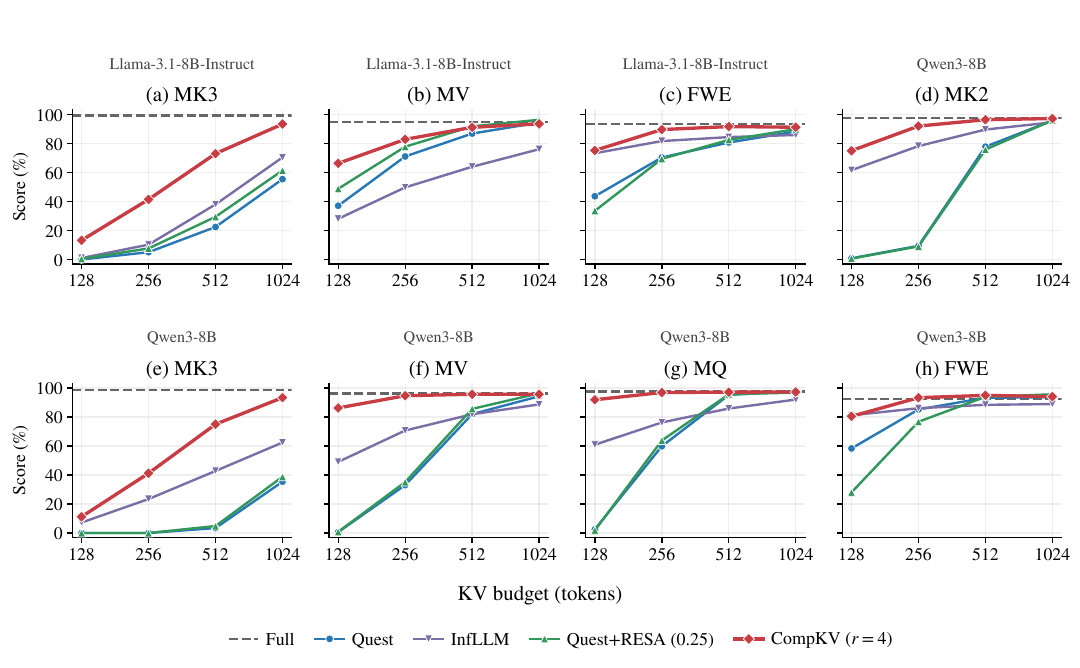}
\caption{Accuracy across KV budgets on selected RULER tasks:
(a--c) Llama-3.1-8B-Instruct and (d--h) Qwen3-8B.
CompKV uses $r=4$. Dashed gray lines denote Full attention.}
\label{fig:budget-accuracy}
\end{figure}

Figure~\ref{fig:budget-accuracy} compares per-step exact-read budgets of 128--1,024 tokens. On MK3, CompKV reaches 73.0\% for Llama and 75.0\% for Qwen with 512 tokens, exceeding the strongest baselines at 1,024 tokens (70.4\% and 62.4\%, respectively). At 128 tokens, Qwen's MV and MQ scores exceed the strongest baseline by 37.2 and 31.0 percentage points, respectively. Several baselines approach Full on FWE, MV, and MQ as budgets increase, narrowing the gaps between methods. 

\section{CPU-Offload Implementation and Detailed Latency Results}

\subsection{Implementation Details}
\label{app:implementation-details}

The CPU-offload adapter stores BF16 KV states and FP32 summaries in pinned CPU memory, while retaining a 16-token active block and reusable workspaces on the GPU. Each append updates the active block's mean keys, mean values, and grouped population variances on the GPU, then copies the current KV states and updated summary to the CPU. Completed blocks' summaries remain unchanged. GPU scoring reuses projected variances for mass estimation and residual scoring, aggregating scores across query heads sharing a KV head. Stable Top-$K$ selection includes one sink block and two recent blocks within the budget and returns block IDs in position order.

The CPU gathers selected KV blocks into a contiguous pinned buffer for one H2D transfer and exact attention with FlashInfer, passing the valid token count explicitly for a partial final block. An auxiliary stream transfers mean values and computes Mean compensation once the selected-block mask is ready, independently of the selected-KV retrieval and exact-attention path. CUDA events synchronize the two branches before merging their contributions under the shared normalization in Equation~\ref{eq:compkv-output}. Intermediate computations use FP32, and the final output is BF16.

\subsection{Detailed Latency Results}
\label{app:latency-details}

Each case runs the pretrained Llama-3.1-8B-Instruct on one NVIDIA H100 80GB HBM3 GPU with four NUMA-local CPU threads and batch size one. Q/K/V and outputs use BF16; summaries and scores use FP32. Following the input construction and collection schedule in Quest's official profiling implementation~\citep{tang2024quest}, we use standard-normal random embeddings matched across methods by a saved RNG state. We prefill $L-256$ tokens and decode 256 steps, defining $L$ as the window endpoint. Steps 193--208 and layers 3--32 yield 480 calls per case. Sparse methods use 16-token blocks with one sink and two recent blocks included in the budget. Full transfers all valid CPU-resident KV at each step. Results are shown in Table~\ref{tab:latency-full}.

A synchronized host timer starts after Q/K/V are ready and stops after the attention output and all state updates complete. It includes KV/summary updates, transfers, selection, CPU gathering, exact attention, compensation, merging, and InfLLM probability maintenance. Input generation, prefill, compilation, projections, RoPE, and MLP are excluded. Primary timing runs without a profiler. We report the mean and sample SD of all 480 calls. Speedups are ratios of reported means.

\begin{table*}[ht]
\centering
\caption{CPU-offload attention-step latency, reported as mean
\textpm{} sample SD (ms) over 480 calls per case. Lengths denote
decode-window endpoints. Full is shared across the three budget rows
at each length. Bold marks the lowest mean in each row.}
\label{tab:latency-full}
\setlength{\tabcolsep}{1.5pt}
\renewcommand{\arraystretch}{1.15}
\begin{tabular*}{\textwidth}{@{\extracolsep{\fill}}ccccccc@{}}
\toprule
Endpoint & Budget & Full & Quest & InfLLM
& CompKV (r=4) & CompKV (r=128) \\
\midrule
\multirow{3}{*}{32K}
& 512 & \multirow{3}{*}{2.536 \textpm{} 0.008}
& 0.595 \textpm{} 0.013 & 0.919 \textpm{} 0.079
& \textbf{0.527} \textpm{} 0.007 & 0.675 \textpm{} 0.006 \\
& 1024 &
& 0.750 \textpm{} 0.013 & 1.084 \textpm{} 0.596
& \textbf{0.665} \textpm{} 0.017 & 0.806 \textpm{} 0.016 \\
& 2048 &
& 0.972 \textpm{} 0.025 & 1.293 \textpm{} 0.051
& \textbf{0.891} \textpm{} 0.024 & 1.038 \textpm{} 0.015 \\
\midrule
\multirow{3}{*}{64K}
& 512 & \multirow{3}{*}{4.995 \textpm{} 0.007}
& 0.944 \textpm{} 0.018 & 1.556 \textpm{} 0.041
& \textbf{0.840} \textpm{} 0.006 & 1.134 \textpm{} 0.010 \\
& 1024 &
& 1.099 \textpm{} 0.016 & 1.706 \textpm{} 0.034
& \textbf{0.916} \textpm{} 0.007 & 1.206 \textpm{} 0.019 \\
& 2048 &
& 1.331 \textpm{} 0.020 & 1.935 \textpm{} 0.056
& \textbf{1.112} \textpm{} 0.021 & 1.391 \textpm{} 0.013 \\
\midrule
\multirow{3}{*}{128K}
& 512 & \multirow{3}{*}{9.966 \textpm{} 0.015}
& 1.625 \textpm{} 0.024 & 2.877 \textpm{} 0.043
& \textbf{1.456} \textpm{} 0.007 & 2.048 \textpm{} 0.033 \\
& 1024 &
& 1.795 \textpm{} 0.020 & 2.996 \textpm{} 0.020
& \textbf{1.536} \textpm{} 0.010 & 2.119 \textpm{} 0.011 \\
& 2048 &
& 2.035 \textpm{} 0.029 & 3.269 \textpm{} 0.056
& \textbf{1.618} \textpm{} 0.018 & 2.200 \textpm{} 0.021 \\
\bottomrule
\end{tabular*}
\end{table*}

\end{document}